\documentclass[10pt,twocolumn,letterpaper]{article}

\usepackage{arXiv}
\usepackage{times}
\usepackage{epsfig}
\usepackage{graphicx}
\graphicspath{{Figures/}}
\usepackage{amsmath}
\usepackage{amssymb}
\usepackage{bm}
\usepackage{mathrsfs}
\usepackage{enumitem}
\usepackage{multirow}
\usepackage{latexsym}
\usepackage{booktabs}

\usepackage{bbding}
\usepackage{tikz}  
\usepackage{pgfplots}
\usetikzlibrary{fit}
\pgfplotsset{compat=newest}%

\usepackage{mathrsfs}
\usepackage{enumitem}
\usepackage{multirow}
\usepackage{latexsym}
\usepackage{booktabs}

\usepackage{array}
\newcolumntype{I}{!{\vrule width 1pt}}
\newcommand*{\Hline}[0]{%
\noalign{\global\setlength{\arrayrulewidth}{1pt}}%
\hline
\noalign{\global\setlength{\arrayrulewidth}{0.4pt}}%
}

\usepackage{colortbl}
\definecolor{mygray}{gray}{.9}
\definecolor{mypink}{rgb}{.99,.91,.95}
\definecolor{mycyan}{cmyk}{.3,0,0,0}

\usepackage[pagebackref=true,breaklinks=true,letterpaper=true,colorlinks,bookmarks=false]{hyperref}

\Finalcopy                      

\begin{document}

\title{NETNet: Neighbor Erasing and Transferring Network for Better \\Single Shot Object Detection}

\author{
Yazhao Li$ ^1$, ~ Yanwei Pang$ ^1$, ~ Jianbing Shen$ ^2$, ~ Jiale Cao$ ^1$, ~ Ling Shao$ ^2$
\\[.152cm]
$ ^1 $Tianjin University, Tianjin, China\\
$ ^2 $Inception Institute of Artificial Intelligence, Abu Dhabi, UAE
}

\maketitle

\begin{abstract}

Due to the advantages of real-time detection and improved performance, single-shot detectors have gained great attention recently. To solve the complex scale variations, single-shot detectors make scale-aware predictions based on multiple pyramid layers. However, the features in the pyramid are not scale-aware enough, which limits the detection performance. Two common problems in single-shot detectors caused by object scale variations can be observed: (1) small objects are easily missed; (2) the salient part of a large object is sometimes detected as an object. With this observation, we propose a new Neighbor Erasing and Transferring (NET) mechanism to reconfigure the pyramid features and explore scale-aware features. In NET, a Neighbor Erasing Module (NEM) is designed to erase the salient features of large objects and emphasize the features of small objects in shallow layers. A Neighbor Transferring Module (NTM) is introduced to transfer the erased features and highlight large objects in deep layers. With this mechanism, a single-shot network called NETNet is constructed for scale-aware object detection. In addition, we propose to aggregate nearest neighboring pyramid features to enhance our NET. NETNet achieves 38.5\% AP at a speed of 27 FPS and 32.0\% AP at a speed of 55 FPS on MS COCO dataset. As a result, NETNet achieves a better trade-off for real-time and accurate object detection.
\end{abstract}


\section{Introduction}

With the emergence of deep neural networks~\cite{alexnet,vgg,resnet}, object detection built on deep networks has achieved significant progress both in detection accuracy~\cite{fastrcnn,RFCN,cornernet} and detection efficiency~\cite{YOLO,YOLOV3}. Beneficial from an optimal trade-off between real-time detection efficiency and accurate detection performance, single-shot detectors~\cite{SSD} have gained increased popularity for various computer vision applications. Despite this success, complex scale variations in practical scenes exist as a fundamental challenge and a bottleneck for accurate object detection~\cite{SNIP,SNIPER,SAN}.

\begin{figure}[t]
    \centering
    \small
\includegraphics[width=8.0cm]{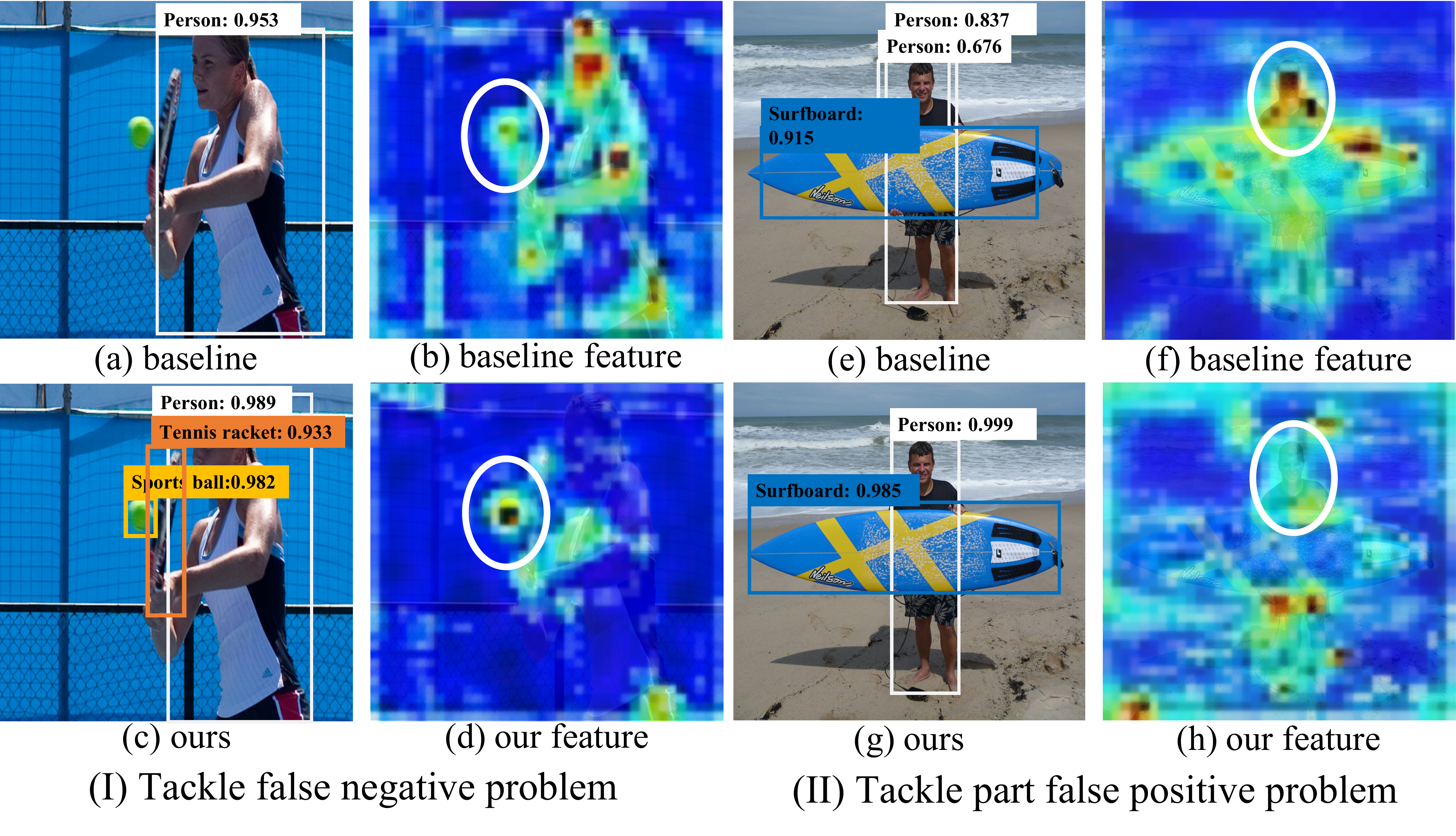}
    \caption{Two common detection problems for baseline SSD~\cite{SSD} and the solution using our NETNet. The visualized features are extracted from the first pyramid layer for detecting small objects. (I) False negative problem. The small objects (tennis racket, sports ball) are missed in (a) because features of small objects are not salient on the corresponding pyramid features (b). Our NETNet can detect small objects with high confidence by erasing the features of large objects and focusing on small objects as (c, d). (II) Part false positive problem. The head is detected as another person in baseline because this part region are highlighted on the features (f) for detecting small objects. Our NETNet can solve this problem by suppressing the salient part features of large objects, as (h).}
    \label{Problem}
\end{figure}

To tackle complex scale variations, the single-shot detector SSD~\cite{SSD} has been proposed and developed based on pyramid feature representation. SSD implements scale-aware object detection by detecting different-sized objects within different layers of the pyramid, which is motivated by the fact that deep-layer features with small feature resolution contain more semantic information for large objects, while the features for small objects are found in the shallow layers with large feature resolution~\cite{RONet,SSDES}. Specifically, shallow layers are responsible for detecting small objects and deep layers are devoted to detecting large objects. Based on feature pyramid, some methods explore to further enhance the feature representation by fusing multi-scale features using an extra feature pyramid, which has proven useful~\cite{Hypernet,FPN, RetinaNet, Reconfig} for improving detection performance. Although single-shot detectors have made great progress for real-time detection and improving detection accuracy by adopting a feature pyramid, several failure cases, such as missing small objects and poor localization~\cite{AOL,CQSSD}, still exist limiting detection performance.

In most previous single-shot detectors, features are scale-confused instead of scale-aware even on one specific pyramid layer. For example, in some shallow layers of a feature pyramid, features for both small and large objects exist. As shown in Fig.~\ref{Problem}, in the shallow features (b) used for detecting small objects, the large-object features dominate the main saliency, weakening the small-object features and thus preventing the detection of small objects (\eg, the sports ball from (a) is not detected in the final result). Additionally, some parts of large objects have strong response regions on shallow features. For example, the head region in Fig.~\ref{Problem}(e) is highlighted in (f), which leads to the wrongly detection of the head region. Thus, the features are scale-confused making it difficult to solve these two problems, \ie, false negative problem and part false positive problem.

With this observation, we propose to generate scale-aware features for better single-shot object detection. To achieve this, redundant features are erased to alleviate feature scale-confusion. Thus, we only keep features of small objects in the shallow layers, erasing features of large objects. Then, we use these small-scale-aware features to detect small objects. As shown in Fig.~\ref{Problem}(d), most of the features of large objects are removed. The features of small objects are thus emphasized, enabling the small sports ball to be detected precisely. The salient features of large objects can also be suppressed to alleviate the part false positive problem, as shown in (h). Meanwhile, transferring these erased features to a suitable scale (\ie, large-scale) space could enhance the features of large objects and improve the overall detection accuracy. 

The main contributions and characteristics of our method are listed as follows:
\begin{itemize}[leftmargin=*]

\item We propose a new Neighbor Erasing and Transferring (NET) mechanism to generate scale-aware features. NET mechanism efficiently reconfigures features between different pyramid layers to alleviate feature scale-confusion. 

\item Two modules, the Neighbor Erasing Module (NEM) and Neighbor Transferring Module (NTM), are designed to unmix the scale confusion and enhance feature aggregation, respectively. The NEM, embedded with a reversed gate-guided erasing procedure, is to extract and erase the large object features from the shallow layers. Then, the large object features are transferred to the deep pyramid layers by the NTM for enhancing the deep features.

\item Based on SSD, a modified single-shot network, NETNet, is constructed by simultaneously embedding the scale-aware features and the scale-aware prediction. In NETNet, we enrich the pyramid features by introducing a Nearest Neighbor Fusion Module (NNFM). 

\item As a result, our NETNet is capable of achieving fast and accurate object detection with a better trade-off than previous single-shot detectors.

\end{itemize}

\begin{figure}
    \centering
    \small
    \includegraphics[width=7.6cm]{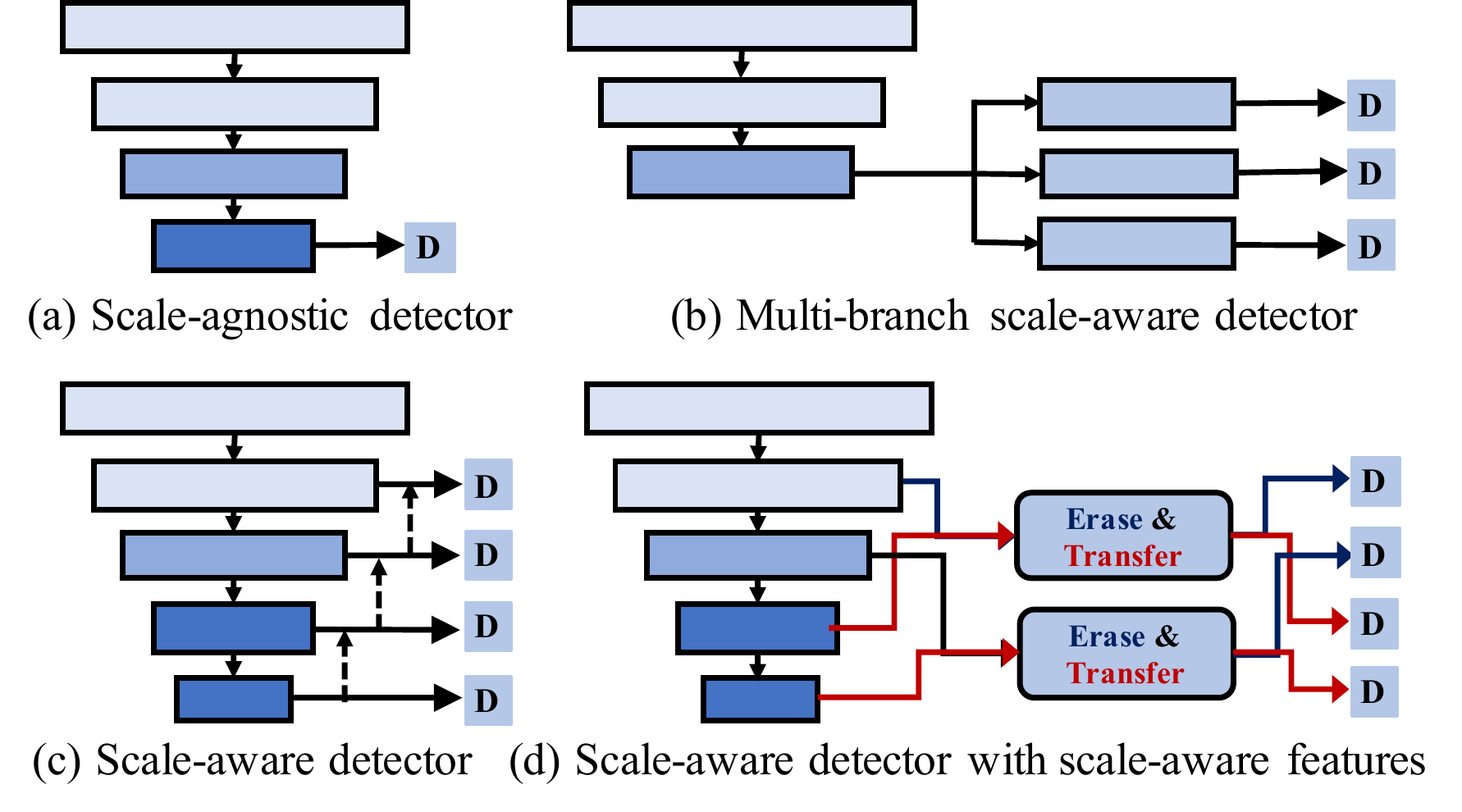}
    \caption{Different detectors for object detection.}
    \label{pyramid}
\end{figure}


\section{Related Work}
\noindent \textbf{Scale-agnostic detectors.} Most recent object detectors are built upon deep networks. The regions with CNN features (R-CNN) methods~\cite{RCNN,fastrcnn} integrate a CNN into object detection and achieve promising performance. As a two-stage method, the Faster R-CNN~\cite{fasterrcnn} proposes a lightweight network for generating proposals and construct the detection network as a complete end-to-end network. Methods like YOLO~\cite{YOLO}, Faster R-CNN~\cite{fastrcnn}, R-FCN~\cite{RFCN}, and other variants~\cite{LHRCNN,Deformable,cascadercnn} have made significant progress for improving detection accuracy and efficiency. As shown in Fig.~\ref{pyramid}(a), this type of methods detect all objects of various scales by utilizing the deepest single-scale high-level features. Thus, these detectors are scale-agnostic detectors.

\noindent \textbf{Scale-aware detectors.} Due to the complex scale variations, many researchers have explored to exploit multi-scale features for improving object detection performance, as shown in Fig.~\ref{pyramid}(b). SSD~\cite{SSD} is a single-shot (\ie, single-stage) detector that proposes to make scale-aware prediction based on multi-layer pyramid features. Features in shallow layers are used for detecting small objects and features in deep layers for large objects. RFBNet~\cite{RFBNet} embeds multi-scale receptive fields to enhance feature discriminability. DES~\cite{SSDES} enriches the semantics of object features through a semantic segmentation branch and a global activation module. FPN~\cite{FPN}, DSSD~\cite{DSSD}, and RONet~\cite{RONet} involve extra top-down feature pyramids and detect objects on each scale of these pyramids as shown in Fig.~\ref{pyramid}(c). Most recent methods~\cite{RetinaNet,PonopFPN,PFPN,STDN,M2DET} have explored the advantages of the pyramid features and have achieved promising results. Kong \etal ~\cite{Reconfig} proposed to reconfigure the pyramid features by aggregating multi-layer features and reassigning them into different levels. Recent TridentNet~\cite{tridentnet} attempts to generate scale-specific features through a parallel multi-branch architectures  as shown in Fig.~\ref{pyramid}(b) by embedding different receptive fields, which achieves promising improvement on two-stage detectors. 

Different from these methods, we propose to generate scale-aware features for single-shot object detection by introducing an erasing and transferring mechanism. The adversarial erasing strategy has also been investigated in weakly supervised object localization ~\cite{eraseORM, eraseACL}, weakly supervised semantic segmentation~\cite{selferasing}, and salient object detection~\cite{reverse}. In these methods, the well recognized regions are erased to refine the prediction results iteratively. Different from them, we propose to reconfigure the pyramid features to scale-aware features by removing the scale-uncorrelated features using an erasing strategy. The erased features in shallow layers are further transferred to enhance the features in deep layers, instead of discarding them as previous erasing methods. As shown in Fig.~\ref{pyramid}(d), we aim to remove the features of large objects from the shallow pyramid layers and generate small-scale-aware features for detecting small objects. The features of large objects in the shallow layers are transferred to enhance the features of deep layers. We then build a single-shot scale-aware detector for more accurate object detection.


\begin{figure*}[t]
    \centering
    \small
    \includegraphics[width=16.5cm]{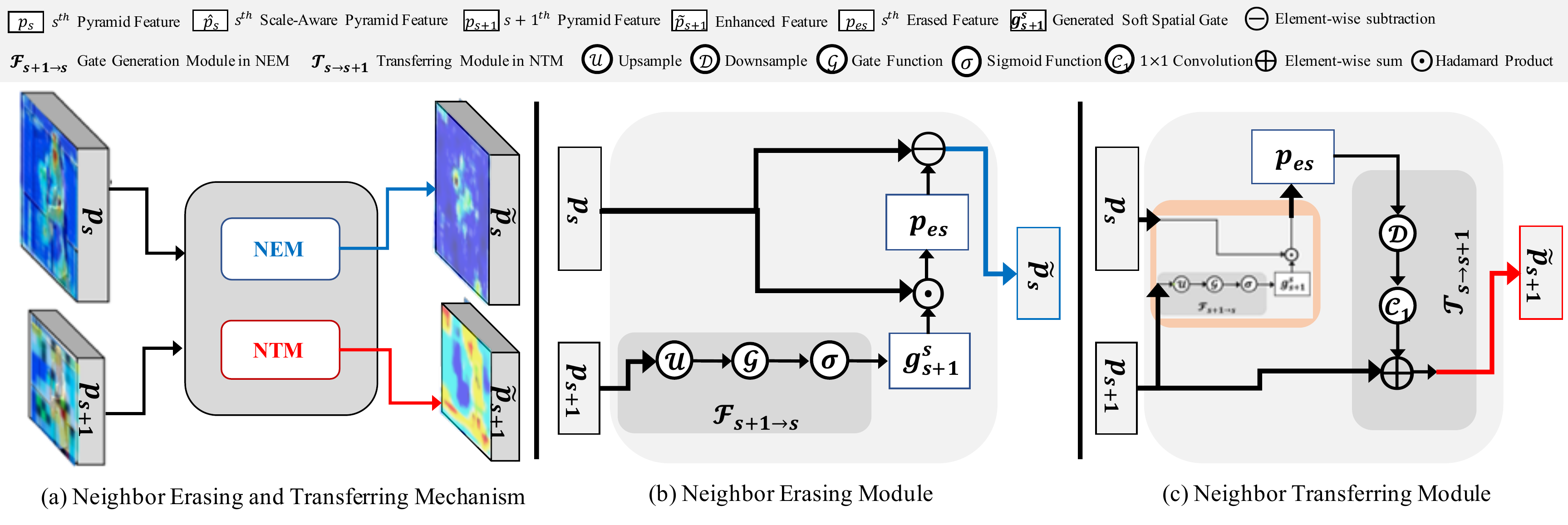}
    \caption{The Neighbor Erasing and Transferring (NET) mechanism (a), with (b) Neighbor Erasing Module (NEM), and (c) Neighbor Transferring Module (NTM). After NETM, $\tilde{p}_s$ highlights small objects, and deep feature $\tilde{p}_{s+1}$ contains more information for larger objects.}  
    \label{NET_NEM_NTM}
\end{figure*}
\section{NET Mechanism} \label{sec:NET}

To tackle complex scale variations, we propose to generate scale-aware features for object detection. As can be observed from Fig.~\ref{Problem}(b) and (f), features in the shallow pyramid layers contain detailed information for both large objects and small objects. However, features for large objects are more salient than small objects, which causes small objects to be missed in Fig.~\ref{Problem}(a) and the part false positive problem in Fig.~\ref{Problem}(e). Instead of promoting feature fusion as previous top-down feature pyramids~\cite{FPN, DSSD}, we propose a NET mechanism to reconfigure the basic pyramid features to scale-aware features for scale-aware object detection. As shown in Fig.~\ref{NET_NEM_NTM}(a), in the NET mechanism, a feature erasing module (\ie, NEM) and a feature transferring module (\ie, NTM) is contained. The NEM is designed to remove large-object features from the shallow layers and emphasize the features of small objects. We then transfer these features using the NTM to enhance the deep features. 

Because our method aims to reconfigure the scale-confused features of the basic pyramid to scale-aware features, we take the typical single-shot detector SSD~\cite{SSD} as our baseline in which a pyramid from the backbone network is adopted for multi-scale prediction.
We first analyze the feature pyramid in the baseline SSD. Then, we present the details of our NEM and NTM in the NET mechanism.

\subsection{Basic Feature Pyramid} \label{subsec:NET_pyramid}

In SSD, a feature pyramid is explored to detect objects with different scales. We denote the objects with a specific scale $s^{th}$ as $x_s$. The objects for all $S$ scales are represented as $X=\left\{x_1, x_2, ..., x_S\right\}$, where $x_1$ represents objects with smallest scale and $x_S$ refers to objects with largest scale.

SSD detects objects in a pyramidal hierarchy by exploiting multiple CNN layers, with each layer is responsible for detecting objects of a specific scale~\cite{EFIP}.
In the feature pyramid with $S$ layers, we denote the features from $s^{th}$ layer as $p_s$ and express all the pyramid features as $P=\left\{p_1, p_2, ..., p_S\right\}$, where $p_1$ represents features with largest resolution in the shallow pyramid layer for detecting small objects $x_1$. With feature pooling in the pyramid, feature resolution is decreased from $p_1$ to $p_S$. Obviously, features for small objects are gradually discarded from shallow to deep layers. Because of the small input image size (\eg, $300\times300$) for SSD, the deep layers (\eg, with spatial size $5\times5$) only contain features for large objects. Thus, we can approximately get:
\begin{equation}
p_s = f_s(x_s,x_{s+1},...,x_S),
\label{eqpy}
\end{equation}
where $f_s(x)$ represents the feature extraction of the pyramid. The feature scale-confusion in a shallow layer (\eg, $p_1$ contains features for various-scale objects) makes detecting small objects difficult and leads to much part detection, as shown in Fig.~\ref{Problem}. We propose to reconfigure the pyramid features to be scale-aware features and solve these problems.

\subsection{Neighbor Erasing Module} \label{subsec:NET_NEM}

To alleviate feature scale-confusion, we propose a Neighbor Erasing Module (NEM) to filter out the redundant features.
Suppose two adjacent pyramid layers, $s^{th}$ layer and $(s+1)^{th}$ layer. Obviously, features in the $s^{th}$ layer $p_s=f_s(x_s,x_{s+1},...,x_S)\in \mathbb{R}^{{h_s}\times {w_s} \times {c_s}}$ have more information for objects $x_{s}$ than features in the $(s+1)^{th}$ layer $p_{s+1}=f_{s+1}(x_{s+1},...,x_S)\in \mathbb{R}^{{h_{s+1}}\times {w_{s+1}} \times {c_{s+1}}}$, where ($h_s> h_{s+1}$, $w_s>w_{s+1}$).
Based on this feature distribution, we can generate features $\tilde{p}_s = f_s(x_s)$ for objects with scale $s$ from the pyramid feature $p_s$, by erasing features $p_{es}=f_s(x_{s+1},...,x_S)$ of objects in a scale range of [$s+1$, $S$] as:
\begin{equation}
     \tilde{p}_s= p_s\ominus p_{es} =f_s(x_s,...,x_S) \ominus f_s(x_{s+1},...,x_S),
    \label{eqerasing} 
\end{equation}
with an element-wise subtraction operation $\ominus$.

Noticing that pyramid feature $p_{s+1}$ only contains information for objects with a scale range of [${s+1}, S$], we therefore use $p_{s+1}$ to guide the feature erasing in Eq.~\ref{eqerasing}. Specifically, we extract the feature $p_{es}$ from $p_s$ by:
\begin{equation}
    p_{es} = p_s \odot \mathcal{F}_{{s+1}\to s}(p_{s+1}),
    \label{eqattentionmal} 
\end{equation}
where $\odot$ refers to Hadamard product. $\mathcal{F}_{{s+1}\to s}(p_{s+1})$ can be represented as a soft spatial gate $g_{s+1}^s\in [0,1]^{h_s\times w_s\times c}$ ($c$ is from $\left\{1,c_s\right\}$). We generate this gate by using the features from the $(s+1)^{th}$ pyramid layer and adopt it to guide suppressing features of objects $(x_{s+1},...,x_S)$ in $p_s$. In our implementation, we calculate this spatial gate as:
\begin{equation}
    g_{s+1}^s = \mathcal{F}_{{s+1}\to s}(p_{s+1}) = \frac{1}{1+e^{-\mathcal{G}(\mathcal{U}(p_{s+1});W_{s+1}^{s})}},
    \label{eqattention} 
\end{equation} 
where $\mathcal{U}(p_{s+1})$ upsamples $p_{s+1}$ to $p_{s+1}^s\in \mathbb{R}^{{h_{s}}\times {w_{s}} \times {c_{s+1}}}$ to keep the consistent spatial resolution between the gate $g_{s+1}^s$ and feature $p_s$. We implement the gate function $\mathcal{G}(.)$ with learnable weights $W_{s+1}^s$. 

In actual, since $\mathcal{G}(.)$ can be represented as a self-attention function~\cite{nonlocal} in which attention for objects can be extracted from the input features, we can construct it based on the spatial attention mechanism in \cite{nonlocal} and \cite{DANet}. Alternately, we can choose to use max pooling or average pooling along channel direction to generate a spatial attention map ($c=1$) like that in~\cite{CBAM} as:
\begin{equation}
    \mathcal{G}(p_{s+1}^s) = \mathcal{P}_{max}(p_{s+1}^s)~~\text{or}~~\mathcal{P}_{avg}(p_{s+1}^s),
    \label{poolingattention} 
\end{equation} 
or combining max pooling $\mathcal{P}_{max}(.)$ and average pooling $\mathcal{P}_{avg}(.)$ by a convolution layer with ${W_{s+1}^{s}}$. In our implementation, we use a $1\times 1 \times c_s$ convolution layer $\mathcal{C}_{1\times 1}$ as:
\begin{equation}
    \mathcal{G}(p_{s+1}^s) = \mathcal{C}_{1\times 1}(p_{s+1}^s;W_{s+1}^{s}),
    \label{eqconvattention} 
\end{equation} 
to generate a channel-wise spatial gate for extracting and suppressing the features of larger objects in $p_s$, since it is proved an optimal trade-off between precision and efficiency as Sec.~\ref{subsec:EXP_abl}. 
In summary, we generate the scale-aware features $\tilde{p}_s$ for smaller objects $x_s$ by suppressing the features of larger objects via a reversed gate as:
\begin{equation}
    \tilde{p}_s = f_s(x_s) = p_s \ominus p_{es} = p_s \ominus (p_s\odot g_{s+1}^{s}).
    \label{eqsummary} 
\end{equation}

\begin{figure*}[t]
    \centering
    \footnotesize
    \begin{center}
    \includegraphics[width=15cm]{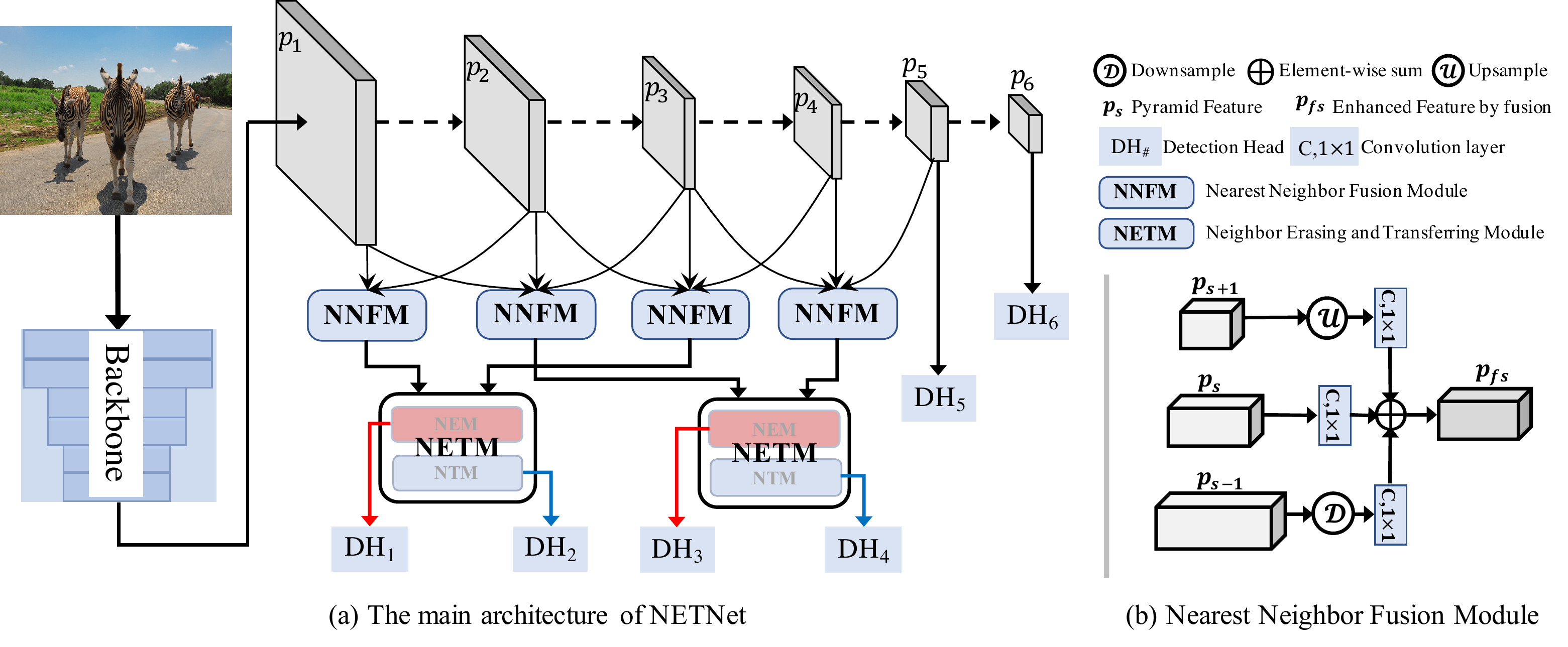}
    \end{center}
    \caption{The proposed NETNet architecture. (a) The main pyramid parts of NETNet. To implement fast object detection, we build a single-shot network based on SSD~\cite{SSD}. We illustrate this architecture by taking the input image with a size of $300\times 300$ as an example. Six pyramid layers are used for building detectors, as in SSD. The embedded NNFM (b) is used for feature fusion before NETM.}
    \label{net2}
\end{figure*}

\subsection{Neighbor Transferring Module}
\label{subsec:NET_NTM}

As discussed above, in the pyramid feature $p_s$, some detailed information (\eg, appearance and edge) for objects $\left\{x_{s+1},x_{s+2},...,x_S\right\}$ is also contained. Although this detailed information disturbs features for detecting smaller objects $x_s$, it is helpful for enhancing the features of larger objects $x_n$ $(n>s)$ for more accurate classification and localization. Therefore, we propose to transfer these features from a shallow layer (\eg, $p_s$) to a deep layer (\eg, $p_{s+1}$). 

As formulated in Section~\ref{subsec:NET_NEM}, the soft spatial gate $g_{s+1}^s\in [0,1]^{h_s\times w_s\times c}$ generated by $p_{s+1}$ has larger activation values on the regions for objects $\left\{x_{s+1},...,x_S\right\}$. Thus, $p_{es}$ in Eq.~\ref{eqattentionmal} helps extract the detailed information of these larger objects.
We then transfer this detailed information $p_{es}$ and obtain the new pyramid features $\tilde{p}_{s+1}\in \mathbb{R}^{{h_{s+1}}\times {w_{s+1}} \times {c_{s+1}}}$ as:
\begin{equation}
\begin{aligned}
    \tilde{p}_{s+1} &= \mathcal{T}_{s\to {s+1}}(p_{es},p_{s+1})\\
    &=\mathcal{C}_{1\times 1}(\mathcal{D}(p_{es});W_s^{s+1})\oplus p_{s+1},
\end{aligned}
\label{eqntm}
\end{equation}
composed of a downsampling operation $\mathcal{D}(.)$ to match the feature resolution and a convolutional layer $\mathcal{C}_{1\times 1}$ with learnable $W_s^{s+1} \in \mathbb{R}^{1\times 1\times {c_{s}} \times {c_{s+1}}}$ to maintain the consistent channel number. We perform an element-wise sum operation $\oplus$ to enhance $p_{s+1}$ by combining the detailed information from $p_{es}$. 
We illustrate this Neighbor Transferring Module (NTM) in Fig.~\ref{NET_NEM_NTM}(c). The enhanced feature $\tilde{p}_{s+1}$ is used as the new pyramid feature for the subsequent scale-aware features generation and scale-aware object detection.


\section{Single-Shot Detector: NETNet} \label{sec:NETNet}

Single-shot object detectors like SSD~\cite{SSD} directly carry out regression and classification based on predefined anchors. This provides the SSD with a better trade-off to achieve real-time detection and promising performance. However, SSD performs poorly for detecting small objects and also suffers from inaccurate localization (\eg, the part detection problem), as shown in Fig.~\ref{Problem}. To solve these problems, we design a new single-shot object detection network, called NETNet embedding the proposed NET mechanism as a scale-aware detector.

In NETNet, we build our backbone network as that of SSD. Taking the network with an input image size 300$\times$300 as an example, we show the main network architecture of NETNet in Fig.~\ref{net2}(a). Features of six pyramid levels $\left\{p_1,p_2,p_3,p_4,p_5,p_6\right\}$ with resolutions $\left\{\right.$38$\times$38, 19$\times$19, 10$\times$10, 5$\times$5, 3$\times$3, 1$\times$1$\left.\right\}$ are extracted from the backbone as the basic feature pyramid. Based on the basic pyramid, we construct our NET Module (NETM) to generate scale-aware features and solve the aforementioned scale problems. In implementation, there are some scale-overlaps~\cite{PANet, couplenet} between the nearest neighbor pyramid levels (\eg, $p_1$ and $p_2$), when configuring the detection anchors and assigning ground truth. Therefore, we build a skipped NETM using our NET mechanism. Additionally, considering that the scale-overlaps make features for one object existing in the nearest neighboring pyramid layers complementary, we introduce a Nearest Neighbor Fusion Module (NNFM) as shown in Fig.~\ref{net2}(b) to enhance the pyramid features firstly by fusing the nearest neighboring pyramid features. Based on the NNFM and NETM, six different detection heads for box regression and classification, are built upon the scale-aware features to construct our scale-aware detector NETNet. We present the details of NETM and NNFM as follows.

\subsection{NETM in a Skip Manner}
\label{subsec:NETNet_NETSkip} 

In typical single-shot detectors, features in the shallow layers (\eg, $p_1$ with larger feature resolution 38$\times$38) are used for detecting smaller objects, while features in deeper layers (\eg, $p_3$ with smaller resolution 10$\times$10) are used for detecting larger objects. Because features with small resolutions (\eg, 3$\times$3) have large receptive fields and less spatial information, we finally embed two NETMs in NETNet for feature erasing and transferring without using features $p_5$ and $p_6$. Due to the anchor configuration in SSD, two anchors in the nearest pyramid layers (\eg, $p_1$ and $p_2$) may share the same ground truth. That is, one small object should be detected in $p_1$ and $p_2$ simultaneously. To avoid disturbing the overlapped supervision, our NETNet is elaborately designed by embedding two skipped NETMs.

One NETM is built upon the pyramid features of $p_1$ and $p_3$. To erase the features of larger objects from the shallow layer $p_1$, we first upsample $p_3$ and use a $1\times 1$ convolution to generate soft spatial gate as Eq.~\ref{eqattention} for larger objects. We evaluate the effects of several different spatial attention methods and choose channel-wise spatial attention as Eq.~\ref{eqconvattention}. Then, an erasing operation in Eq.~\ref{eqsummary} generates features for smaller objects. We also embed a light fusion module into NETM to make the generated scale-aware features more robust. The fusion module is constructed as a residual block as in~\cite{resnet} by stacking ($1\times 1$ convolution, $3\times 3$ convolution, and $1\times 1$ convolution) with a skip connection. When applying the transferring module NTM, we first acquire the detailed information $p_{es}$ that is helpful for larger objects from $p_1$ as Eq.~\ref{eqattentionmal}. Then, this detailed information enhances the features $p_3$ as Eq.~\ref{eqntm}. The other NETM is built upon pyramid features of $p_2$ and $p_4$ with the similar configuration.

\subsection{Nearest Neighbor Fusion Module} 
\label{subsec:NETNet_NNFM}

As pointed out in feature pyramid studies~\cite{PFPN, EFIP}, features from neighboring pyramid layers are complementary. Thus, incorporating context information from different layers promotes feature representation. Combining features from top to bottom is typically done to build a feature pyramid~\cite{DSSD}. However, since our purpose is to remove large-object features from the shallow layers and generate scale-aware features, introducing other more scale features may increase the feature scale-confusion problem. Therefore, we propose a more effective fusion module, NNFM, to enhance the pyramid features. 

As shown in Fig.~\ref{net2}(b), in NNFM, only features from the adjacent pyramid layers are fused as: 
\begin{equation}
    p_{fs} = \mathcal{H}_{s-1}(p_{s-1})\oplus \mathcal{H}_{s}(p_{s})\oplus \mathcal{H}_{s+1}(p_{s+1}),
\end{equation}
where we denote the fused features of $s^{th}$ pyramid layer as $p_{fs} \in \mathbb{R}^{{h_{s}}\times {w_{s}} \times {c_{s}}}$. $\mathcal{H}_{s-1}$ is constructed by a pooling layer and a $1\times 1$ convolutional layer. $\mathcal{H}_{s}$ is constructed by a $1\times 1$ convolutional layer. $\mathcal{H}_{s+1}$ is constructed by a bilinear upsampling layer and a $1\times 1$ convolutional layer. Finally, these features are fused by an element-wise sum operation. Thus, we enhance the $p_2$ features by aggregating complementary information from $p_1$, $p_2$, and $p_3$, instead of using the features $\left\{ p6,p5,p4,p3,p2\right\}$ like a top-down pyramid network. Performing NNFM will not aggravate the feature scale-confusion, since the information of tiny objects from $p_1$ is discarded using pooling operation and the information of larger objects from $p_3$ will be erased by the subsequent NEM. As a result, the features of objects which should be detected on $p_2$, are enhanced by fusing the complementary information with NNFM.

\section{Experiments}
\label{sec:EXP}

\noindent \textbf{Dataset:} We evaluate our method on the benchmark detection dataset, MS COCO~\cite{COCO} dataset (\ie, COCO). It has 80 object categories and more than 140k images. Following~\cite{SSD,FPN}, we train our NETNet on the union (\textit{trainval35k}) of 80k training images and a 35k subset of validation images, and conduct ablation evaluations on the remaining 5k validation images (\textit{minival}). The final results are obtained by testing on the 20k test images (\textit{test-dev}) and submitted to the official server. The variations in scale of objects in COCO are complex. 
AP$_s$, AP$_m$, and AP$_l$ evaluate the detection precision for three scales of objects. 

\noindent \textbf{Training protocols:} We re-implement the SSD~\cite{SSD} as our baseline based on a Pytorch framework. All the models are trained over 160 epochs with the same training loss as SSD. For ablation experiments, we set the initial learning rate as 0.002 and decrease it by a factor of 0.1 after the 90$^{th}$, 120$^{th}$, and 140$^{th}$ epochs, respectively. We follow~\cite{RFBNet}, using a warm-up learning rate in the first 5 epochs. We set the weight decay to 0.0005 and the momentum to 0.9. Each model is trained with a batch size of 32 on 2 GPUs. Results are reported using the standard COCO-style metric. 

\begin{table}[t]
\centering
\small
\setlength{\tabcolsep}{1.9mm}
\begin{center}
\renewcommand\arraystretch{1.0}
\begin{tabular}{r||ccc|ccc}
\Hline
     Methods & AP  &AP$_{50}$ &AP$_{75}$ &AP$_s$ &AP$_m$ &AP$_l$ \\ \hline \hline
     Baseline SSD      &25.1	&41.8	&26.1	&6.3	&28.3	&43.3		\\ \hline
     NEM     &29.4	&48.9	&30.4	&13.2	&32.2	&44.3		\\ 
     NTM     &25.8	&42.4	&26.9	&6.5	&28.5	&44.4		 \\ 
     NETM   &30.4	&49.7	&31.4   &13.4	&33.0   &45.6	\\ \hline
     NETM + TDP  &30.6	&49.9	&31.9  &12.8	&33.0	&\textbf{46.3}	\\\hline 
     \textbf{NETNet}    &\textbf{31.1}	&\textbf{50.5}	&\textbf{32.4}	&\textbf{13.6}	&\textbf{35.0}	&45.4\\ \hline
\end{tabular}
\end{center}
\caption{Ablation evaluation for NETM and NNFM on the MS COCO \textit{minival} set. NETNet is our model with NETM and NNFM.}
\label{NET}
\end{table}

\begin{table}[t]
\centering
\small
\setlength{\tabcolsep}{1.7mm}
\begin{center}
\renewcommand\arraystretch{1.1}
\begin{tabular}{r||ccc|ccc}
\Hline
 Methods & AP  &AP$_{50}$  &AP$_{75}$ &AP$_s$ &AP$_m$ &AP$_l$ \\ \hline \hline
     Max Attention     &28.7	&47.3	&29.9	&11.5	&31.4	&43.4	\\
     Mean Attention    &28.8	&47.6	&29.6	&12.5	&32.0	&43.9	\\
     Global Attention     &29.3	&48.6	&\textbf{30.5}	&12.5	&32.0	&44.2	\\ \Hline
     \textbf{NEM}    &\textbf{29.4}	&\textbf{48.9}	&{30.4}	&\textbf{13.2}	&\textbf{32.2}	&\textbf{44.3}	\\ \hline
\end{tabular}
\end{center}
\caption{Ablation evaluation for different attentions of NEM.} 
\label{attention}
\end{table}

\subsection{Ablation Study} 
\label{subsec:EXP_abl}  
\noindent \textbf{Configuration of NETNet.}
For ablation experiments, we construct NETNet with a VGG-16 backbone pretrained on ImageNet~\cite{imagenet}, and train the models with an input size of 300$\times$300. Following SSD, we truncate the final fully connected layers of the backbone and add a series of smaller convolutional layers to construct the feature pyramid.

\noindent \textbf{Evaluation of NETNet:}

\textit{Overall NEM.} As shown in Table~\ref{NET}, compared with SSD, NEM yields a large margin of absolute improvement of 4.3\% AP. Because our NEM can remove the features of larger objects from the shallow layer to solve feature confusion, the salient regions can be suppressed and features for smaller objects can be activated to improve the performance for detecting smaller objects. We obtain a 6.9\% AP improvement for small objects and 3.9\% AP improvement for medium objects, which demonstrates the effectiveness of NEM for feature erasing. 

\textit{NTM and NETM.} We propose to transfer features using NTM to complement the detailed information of larger objects. As shown in Table~\ref{NET}, using only NTM brings a 1.1\% improvement for large objects because of the enhanced features for large objects. Combining NEM and NTM promotes each module to learn better features through an adversarial strategy. Our NETM using NEM and NTM further improves the overall AP by 1.0\%.

\textit{NNFM.} We compare our NNFM for feature fusion with a typical Top-Down Pyramid (TDP) like FPN~\cite{FPN} based on our NETM. When combing the TDP with our NETM, a slight overall improvement, 0.2\% AP, is achieved. However, we find the detection performance for small objects degrades by using TDP (from 13.4\% AP to 12.8\% AP), which may be caused by the feature confusion that is not consistent with our NET mechanism. When combining the NETM with NNFM (\ie, NETNet), a 31.1\% AP performance is obtained. Our NNFM further improves the performance for medium objects by a large margin (2.0\%).

\noindent \textbf{Evaluation of NEM:}

\textit{Attention for NEM.} We train our network with only two NEMs to evaluate different spatial gate generation methods as discussed in Sec.~\ref{subsec:NET_NEM}. Due to the large computation consumption of spatial attention method in~\cite{nonlocal, DANet}, we only implement a simplified one as 'Global Attention' by reducing the inner channel number. 'Mix' represents combining 'Max' and 'Avg' attention. As presented in Table~\ref{attention}, using attention as Eq.~\ref{eqconvattention} in our NEM, which generates a channel-wise spatial gate for each channel of the shallow pyramid features, obtains a better performance of 29.4\% AP. We visualize some examples in supplementary material.

\begin{table}[t]
\centering
\small
\setlength{\tabcolsep}{2.0mm}
\begin{center}
\renewcommand\arraystretch{1.0}
\begin{tabular}{r||ccc|ccc}
\Hline
     Methods & AP  &AP$_{50}$ &AP$_{75}$ &AP$_s$ &AP$_m$ &AP$_l$ \\ \hline \hline
     Baseline SSD      &25.1	&41.8	&26.1	&6.3	&28.3	&43.3\\ \hline
     NEM$_{13}$     &28.9	&48.7	&30.2	&12.8	&31.0	&44.4\\
     NEM$_{24}$      &28.5	&46.6	&30.0	&10.6	&31.7	&44.5\\  \hline
     NNEM &29.1	&48.8	&30.1	&12.7	&31.9	&{44.4}\\ \hline 
     \textbf{NEM}     &\textbf{29.4}	&\textbf{48.9}	&\textbf{30.4}	&\textbf{13.2}	&\textbf{32.2}	&44.3\\ \hline
\end{tabular}
\end{center}
\caption{Ablation evaluation for different NEMs on \textit{minival} set.} 
\label{2NE}
\end{table}


\begin{table*}[t]
\footnotesize
\centering
\begin{center}
\renewcommand\arraystretch{1.0}
\setlength{\tabcolsep}{3.0mm}
\begin{tabular}{r|cc|cc|c|cc|ccc}
\Hline
      Methods & Backbone & Image Size & Time (ms) & FPS & AP  &AP$_{50}$ &AP$_{75}$ &AP$_s$ &AP$_m$ &AP$_l$ \\ \hline \hline
     \textbf{Two-stage detectors:}&&&&&&&&&&\\
     Faster~\cite{fasterrcnn}	           &VGG-16	       &1000$\times$600	     &147	  & 6.8	    &24.2	&45.3	&23.5	&7.7	&26.4	&37.1\\
     Faster-FPN ~\cite{FPN}	       &ResNet-101	   &1000$\times$600	     &190	  & 5.3   	&36.2	&59.1	&39.0	&18.2	&39.0	&48.2\\
     R-FCN~\cite{RFCN}	           &ResNet-101	   &1000$\times$600	     &110	  & 9.1   	&29.9	&51.9	&- 	    &10.8	&32.8	&45.0\\
     
     CoupleNet~\cite{couplenet}	       &ResNet-101	   &1000$\times$600	     &120	  & 8.0  	&34.4	&54.8	&37.2	&13.4	&38.1	&50.8 \\
     Mask R-CNN~\cite{maskrcnn}	       &ResNext-101	   &1280$\times$800	     &210	  & 4.8	    &39.8	&62.3	&43.4	&22.1	&43.2	&51.2\\
     Cascade R-CNN~\cite{cascadercnn}     &Res101-FPN     &1280$\times$800      &141     & 7.1     &42.8   &62.1   &46.3   &23.7   &45.5   &55.2\\ \hline \hline
     
\textbf{Anchor-free detectors}:& & & & & & & & & &\\
CornerNet~\cite{cornernet} &Hourglass-104 &511$\times$511 &244 &4.1 &40.5 &56.5 &43.1 &19.4 &42.7 &53.9\\
     CenterNet~\cite{centernet} &Hourglass-104 &511$\times$511 &340 &2.9 &44.9 &62.4 &48.1 &25.6 &47.4 &57.4 \\
     FCOS~\cite{FCOS} &Res101-FPN &1333$\times$800 &- &- &41.5 &60.7 &45.0 &24.4 &44.8 &51.6\\ \hline \hline
     \textbf{Single-stage detectors:} &&&&&&&&&&\\ 
     SSD300~\cite{SSD}	           &VGG-16	       &300$\times$300	     & 17*	  & 58.9    	&25.1	&43.1	&25.8	&6.6	&25.9	&41.4\\
     DFPR~\cite{Reconfig}	           &VGG-16	       &300$\times$300	     & -      & -           &28.4	&48.2	&29.1    &-       &-    &-   \\
     PFPNet-S300~\cite{PFPN}	   &VGG-16	       &300$\times$300       & -      & -       	&29.6	&49.6	&31.1	&10.6	&32.0	&44.9\\
     RefineDet320~\cite{RefineDet}	   &VGG-16	       &320$\times$320	     &26	  & 38.7        	&29.4	&49.2	&31.3	&10.0	&32.0	&44.4\\
     RFBNet~\cite{RFBNet}	           &VGG-16	       &300$\times$300	     & 15 (19*)	  & 66.7    	&30.3	&49.3	&31.8	&11.8	&31.9	&45.9\\
     EFIP~\cite{EFIP}  &VGG-16   &300$\times$300  & 14 	  & 71.4  &30.0 &48.8 &31.7 &10.9 &32.8 &46.3\\
     HSD~\cite{HSD} &VGG-16 &320$\times$320 &25 &40.0 &33.5 &53.2 &36.1 &15.0 &35.0 &47.8\\
     
\rowcolor{mygray}      NETNet (ours)	           &VGG-16	       &300$\times$300	     & 18	  & 55.6    	&{32.0}	&{51.5}	&{33.6}	&{13.9}	&{34.5}	&46.2\\
\rowcolor{mygray}      NETNet+Ref~\cite{HSD}	           &VGG-16	       &320$\times$320	     &-	  &-    	&{34.9}	&{53.8}	&{37.8}	&{16.3}	&{37.7}	&48.2\\\hline
     DSSD513~\cite{DSSD}	       &ResNet-101	   &513$\times$513	     & 182    & 5.5         &33.2	&53.3	&35.2	&13.0	&35.4	&51.1\\
     RetinaNet~\cite{RetinaNet}	       &ResNet-101	   &500$\times$500	     & 90	  & 11.1    	&34.4	&53.1	&36.8	&14.7	&38.5	&48.5\\

     STDN512~\cite{STDN}	       &DenseNet-169   &513$\times$513	     & -	  &  -       	&31.8	&51.0	&33.6	&14.4	&36.1	&43.4\\
     
     DFPR~\cite{Reconfig}	           &ResNet-101	   &512$\times$512	     & -      &  -          &34.6	&54.3	&37.3	&14.7	&38.1	&51.9\\

     RefineDet512~\cite{RefineDet}	   &ResNet-101	   &512$\times$512	     & -	  &  -       	&36.4	&57.5	&39.5	&16.6	&39.9	&51.4\\
     SSD512~\cite{SSD}	           &VGG-16	       &512$\times$512	     & 28	  & 35.7    	&28.8	&48.5	&30.3	&10.9	&31.8	&43.5\\
     DES512~\cite{SSDES}	           &VGG-16	       &512$\times$512	     & -      &  -      	&32.8	&53.2	&34.6	&13.9	&36.0	&47.6\\
     RFBNet~\cite{RFBNet}	           &VGG-16	       &512$\times$512	     & 33 (37*)	  & 30.3    	&34.4	&55.7	&36.4	&17.6	&37.0	&47.6\\
     EFIP~\cite{EFIP} &VGG-16	       &512$\times$512 &29 &34.5 &34.6 &55.8 &36.8 &18.3 &38.2 &47.1\\

     TripleNet~\cite{triplenet} &ResNet-101 &512$\times$512 &- &- &37.4 &59.3 &39.6 &18.5 &39.0 &52.7\\

\rowcolor{mygray}     NETNet (ours)           &VGG-16   	   &512$\times$512	     & 33	  & 30.3    	&36.7	&57.4	&39.2	&{20.2}	&39.2	&49.0\\

\rowcolor{mygray}    NETNet (ours)	           &ResNet-101	   &512$\times$512	     & 37	  & 27.0    	&{38.5}	&{58.6}	&{41.3}	&19.0	&{42.3}	&{53.9}\\\hline

\end{tabular}
\end{center}
\caption{Comparison on the MS COCO \textit{test-dev} set. The results are reported for the case of single-scale inference. We test the time on a Titan X Pascal GPU with Pytorch 0.3.1. Times with * are obtained by testing in the same environment with NETNet.}
\label{Test}
\end{table*}


\textit{NEM on different layers.} We evaluate the influence of each NEM and show the results in Table~\ref{2NE}. By only adding NEM on $p_1$ and $p_3$, we obtain a 6.5\% AP improvement (NEM$_{13}$) on AP$_s$, which is better than that of NEM$_{24}$ (on $p_2$ and $p_4$) because there are more small objects features in $p_1$. We obtain a better improvement for medium objects by NEM$_{24}$. There is some ground truth and feature overlap in $p_1$ and $p_2$, which yields the improvements for both small and medium objects using each NEM. We obtain the best result by combining them. These results demonstrate the effectiveness of our method for erasing redundant features.

\textit{Skipped NEM.} We also construct a model by adding three regular NEMs built upon ($p_1$, $p_2$), ($p_2$, $p_3$), and ($p_3$, $p_4$), respectively.  This is a type of nearest neighbor erasing module built upon the features of two nearest neighbor layers. We denote this model as NNEM in Table~\ref{2NE}. The NNEM model obtains a lower performance (29.1\%) than our NEM (29.4\%). Because the same ground truth may be assigned to predefined anchors from two neighboring layers, using NNEM disturbs the ground truth supervision. Using the skipped NEM helps the network achieve better results for detecting small objects and medium objects.

\noindent \textbf{Evaluation of network configurations:} 
We evaluate the performance of NETNet with different configurations. By refining the learning rate (using 0.004 as the initial learning rate), we achieve a final best performance of 31.8\% AP with a 300$\times$300 input size. When we further use the refined prediction procedure in ~\cite{HSD}, a 34.7\% AP performance is obtained. In addition, larger image size and better backbone help improve the performance. With VGG-16 and a 512$\times$512 size, 36.1\% AP is obtained. Using ResNet-101 brings NETNet to a top performance, 38.2\% AP.


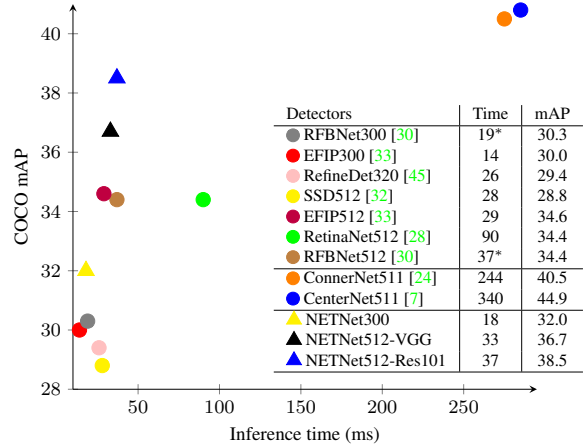
\begin{figure}[t]
\begin{center}
\footnotesize
\begin{tikzpicture}[scale=0.9] 
 \begin{axis}[
 axis lines = left,
 ymin=28, ymax=41, 
 xmin=10, xmax=295,
 xlabel=Inference time (ms),
 ylabel= \footnotesize{COCO mAP}]
 \coordinate (legend) at (axis description cs:0.99,0.006);
  \addplot[only marks,
      mark=otimes*, yellow,
      mark size=3.0pt
      ]
      coordinates {
      (28,28.8)};\label{plot:ssd}

      
      \addplot[only marks,
      mark=otimes*, green,
      mark size=3.0pt
      ]
      coordinates {
      (90,34.4)};\label{plot:retinanet}

      \addplot[only marks,
      mark=otimes*, brown,
      mark size=3.0pt
      ]
      coordinates {
      (37,34.4)};\label{plot:RFBNet}
      
      \addplot[only marks,
      mark=otimes*, pink,
      mark size=3.0pt
      ]
      coordinates {
      (26,29.4)};\label{plot:RefineDet320}
      
      \addplot[only marks,
      mark=otimes*, red,
      mark size=3.0pt
      ]
      coordinates {
      (14,30.0)};\label{plot:EFIP300}

      \addplot[only marks,
      mark=otimes*, purple,
      mark size=3.0pt
      ]
      coordinates {
      (29,34.6)};\label{plot:EFIP}
      
      \addplot[only marks,
      mark=otimes*, gray,
      mark size=3.0pt
      ]
      coordinates {
      (19,30.3)};\label{plot:RFBNet300}
      
      
      \addplot[only marks,
      mark=triangle*, yellow,
      mark size=4.0pt
      ]
      coordinates {
      (18,32.0)};\label{plot:ours300}
      
      \addplot[only marks,
      mark=triangle*, black,
      mark size=4.0pt
      ]
      coordinates {
      (33,36.7)};\label{plot:oursvgg}
      
      \addplot[only marks,
      mark=triangle*, blue,
      mark size=4.0pt
      ]
      coordinates {
      (37,38.5)};\label{plot:oursres}
      
      \addplot[only marks,
      mark=otimes*, orange,
      mark size=3.0pt
      ]
      coordinates {
      (275,40.5)};\label{plot:connernet}

      \addplot[only marks,
      mark=otimes*, blue,
      mark size=3.0pt
      ]
      coordinates {
      (285,40.8)};\label{plot:centernet}
      
 \end{axis}
  \node[draw=none,fill=none,anchor= south east] at (legend){\resizebox{0.50\linewidth}{!}{ 
  		\begin{tabular}{l|c|c}
        Detectors  & Time &mAP \\ \hline
        \ref{plot:RFBNet300} RFBNet300~\cite{RFBNet}& 19$^*$  & 30.3 \\
        \ref{plot:EFIP300} EFIP300~\cite{EFIP}& 14   & 30.0 \\
        \ref{plot:RefineDet320} RefineDet320~\cite{RefineDet} &26 &29.4 \\
        \ref{plot:ssd} SSD512~\cite{SSD}& {28}   & 28.8  \\

        \ref{plot:EFIP} EFIP512~\cite{EFIP}& 29   & 34.6 \\
        \ref{plot:retinanet} RetinaNet512~\cite{RetinaNet}& 90   & 34.4 \\
        \ref{plot:RFBNet} RFBNet512~\cite{RFBNet}& 37$^*$  & 34.4 \\\hline
        \ref{plot:connernet} ConnerNet511~\cite{cornernet}& 244  & 40.5 \\
        \ref{plot:centernet} CenterNet511~\cite{centernet}& 340  & 44.9 \\
        
        
        \hline
        \ref{plot:ours300} NETNet300 & 18   &32.0   \\
        \ref{plot:oursvgg} NETNet512-VGG &33   &36.7 \\
        \ref{plot:oursres} NETNet512-Res101 & 37   &38.5  \\
        \hline
        \end{tabular}}};
     \end{tikzpicture}
  \end{center}
    \caption{Accuracy (mAP) vs. speed (ms) comparison. Methods in the top-left corner have better overall performance.
  } \label{figuretimevs} 
  \end{figure}

\subsection{Results on COCO Test Set}
We evaluate NETNet on the COCO \textit{test-dev} set and compare it with previous state-of-the-art methods, as shown in Table~\ref{Test}. Our NETNet outperforms the baseline SSD significantly with only a slight extra time cost. With an input size of 300$\times$300 and VGG-16, our NETNet obtains 32.0\% AP with 55.6 FPS, which outperforms other state-of-the-art single-shot detectors with a similar configuration. Employing the refinement in~\cite{HSD} helps NETNet obtain a top performance 34.9\% AP. When testing with an image size of 512$\times$512, NETNet obtains 36.7\% (30.3 FPS) with VGG-16 and 38.5\% (27.0 FPS) with ResNet-101. Some anchor-free methods achieve better detection accuracy, but they are generally require more than 100 ms to process one image. As shown in Fig.~\ref{figuretimevs}, our method achieves an optimal trade-off for accurate detection while maintaining a fast speed.


\begin{figure}
    \centering
    \includegraphics[width=8.0cm]{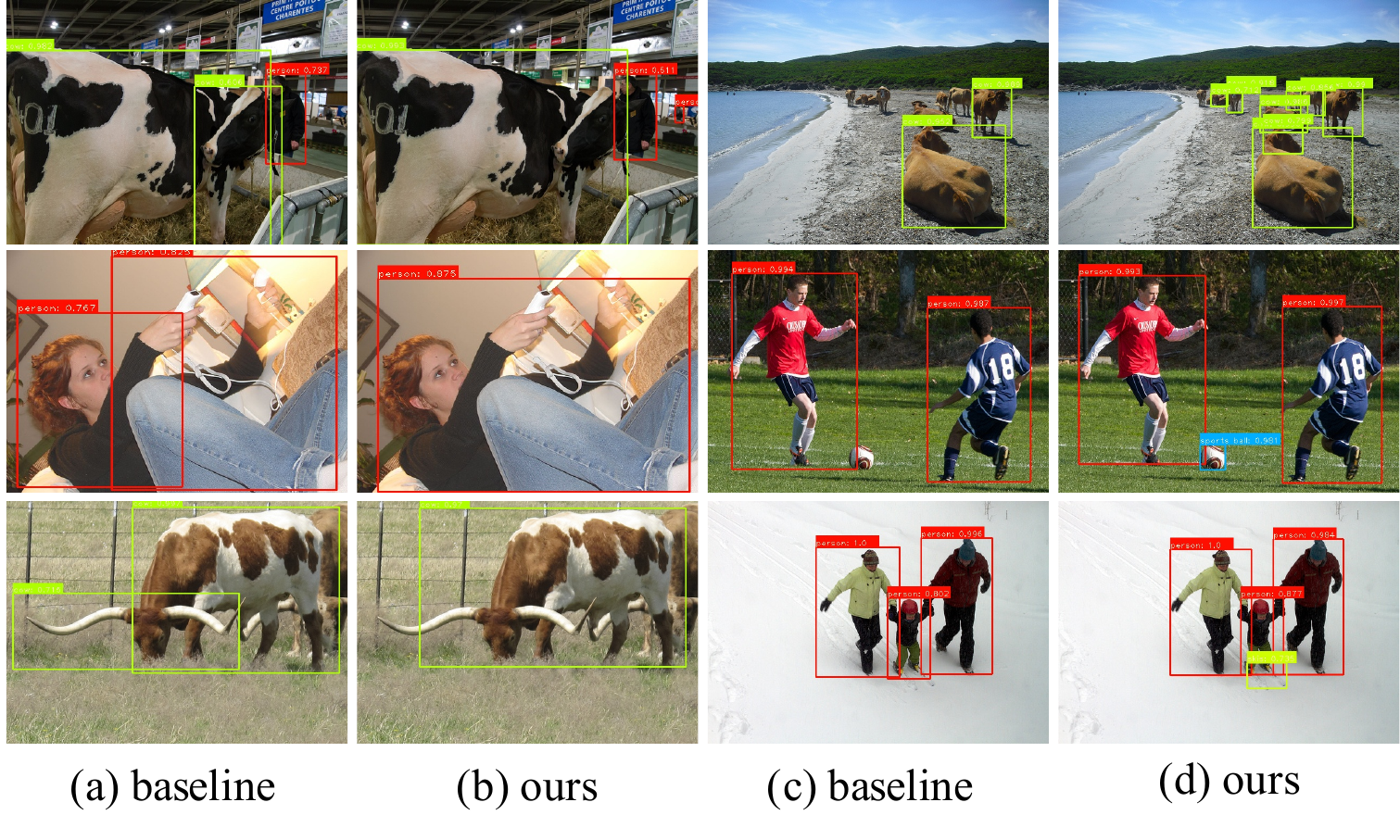}
    \caption{Detection results visualization. Our method can alleviate the part false positive problem as (b) and the small objects missing (false negative problem) as (d). More detection results can be found in the supplementary material.}
    \label{figurevis}
\end{figure}

\section{Discussion}

Different from previous pyramid methods, NET mechanism helps reconfigure the basic pyramid to be scale-aware features which are more suitable for scale-aware detection. In another side, because we need use shallow features to generate deep features by progressively convolution operations in a network, using a direct hard supervision will force the large object regions in shallow layers of the backbone to be background and harm the feature learning of deep layers. NET works like a soft supervision by introducing a reversed feedback from high-level features for feature erasing, which will not harm the feature learning but enhance the information aggregation in the backbone pyramid. More visualization analysis can be found in the supplementary material.

In addition, we carry out an error analysis to further demonstrate the effectiveness of our method for solving the false positive (FP) problem and false negative problem (FN,~\ie, missing detection). For fair comparison, we use the detection results on the \textit{minival} set by SSD and NETNet (31.8\% AP) with VGG-16 and 300$\times$300 image size. 

\begin{figure}
    \centering
     \includegraphics[width=8.0cm]{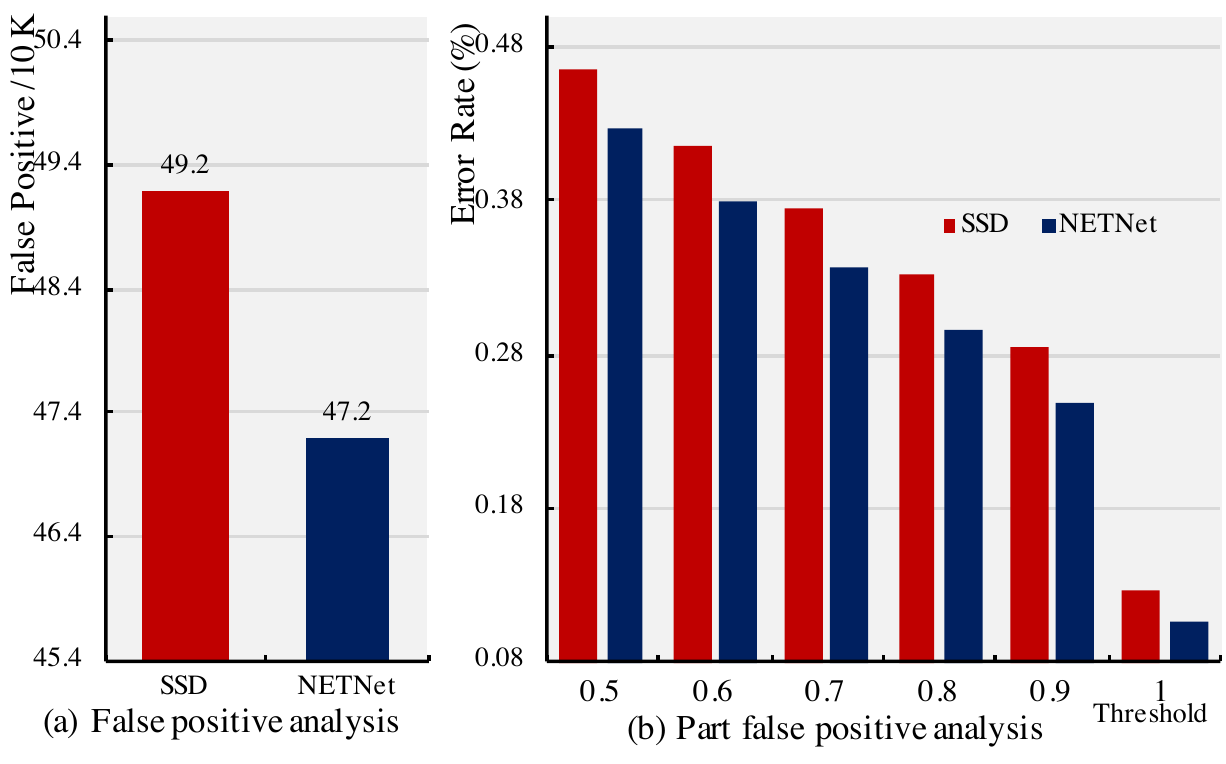}

    \caption{Error analysis for total false positive problem (a) and part false positive problem (b) on the MS COCO \textit{minival} set.}
\label{plterror}
\end{figure}

\noindent \textbf{Tackling FP problem.} By treating the predicted box, which has a IoU \textless 0.5 with the ground truth as a FP sample, we conduct a statistical analysis for the FP problem. In total, there are about 20k less FP samples by our method than SSD as shown in Fig.~\ref{plterror}(a), which demonstrates our method can alleviate this problem. We further analyze the part false positive (PFP) problem based on the PFP samples under different thresholds. The part rate $p_\theta$ is calculated as the ratio of intersection region (between one predicted FP box and the ground truth) over the area of the predicted box. If $p_\theta$ is higher than the threshold, the FP box is regarded as a PFP sample. We present the PFP error in Fig.~\ref{plterror}(b). The x-axis denotes the thresholds and y-axis represents the ratio of PFP sample number over total predicted box number. Our method can reduce the PFP error. We visualize some detection results in Fig.~\ref{figurevis} (a) and (b). 

\begin{figure}
\small
    \centering
    \setlength{\tabcolsep}{2mm}
    \begin{tabular}{cccc}
     \includegraphics[width=3.8cm]{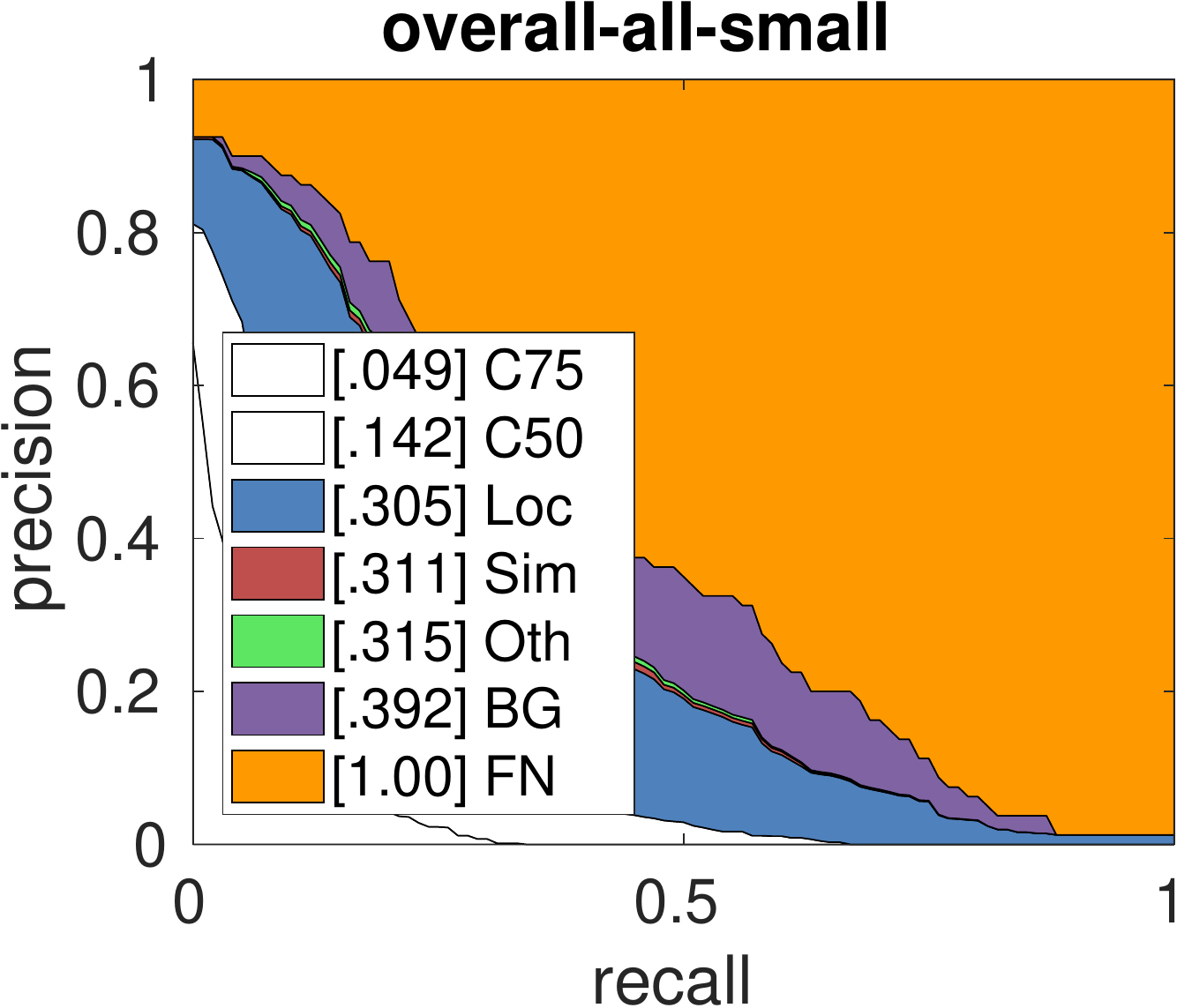}&  \includegraphics[width=3.8cm]{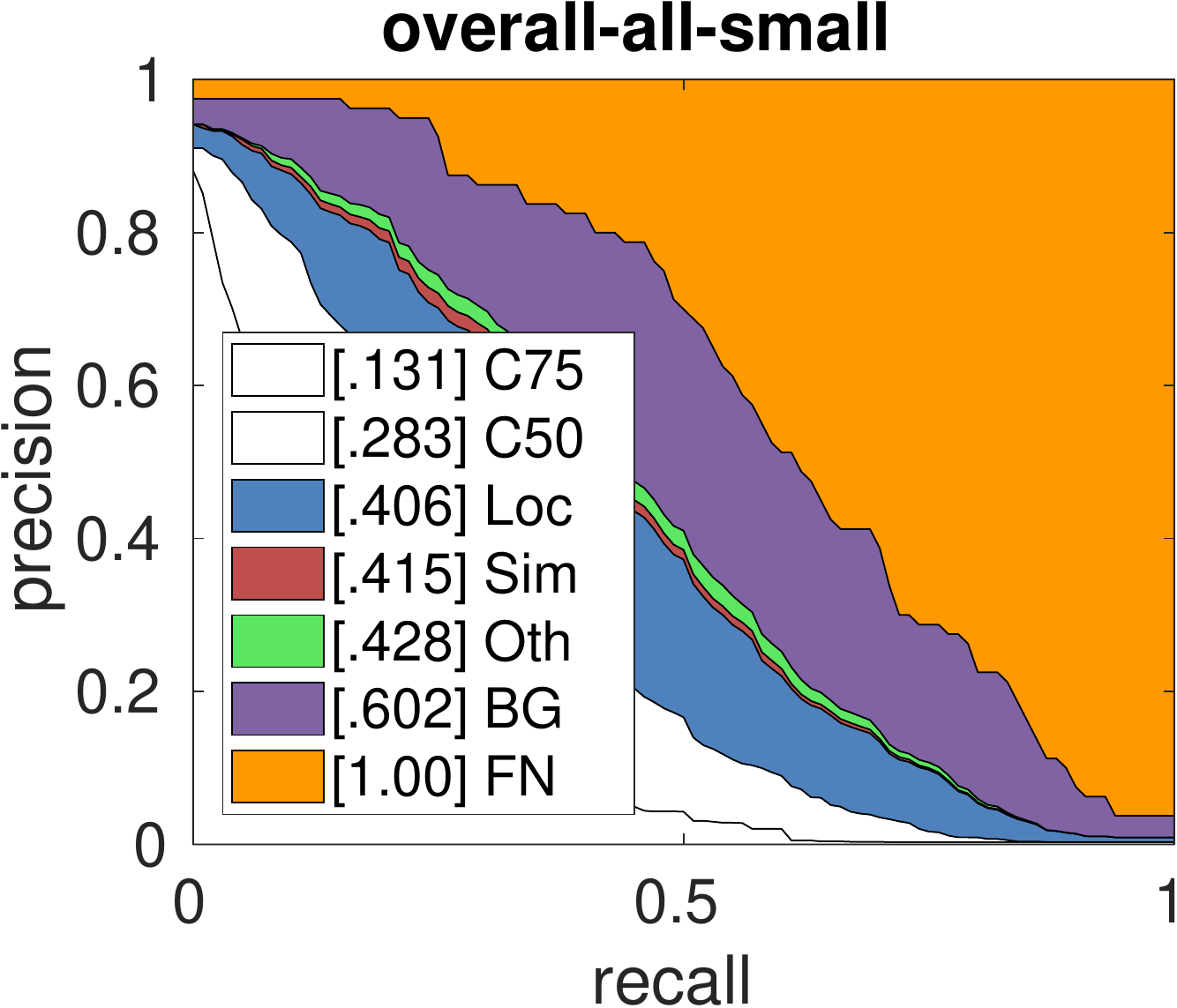}\\
    (a) \scriptsize{SSD} &(b) \scriptsize{NETNet} 
    \end{tabular}
    \caption{Error analysis for (a) baseline SSD and (b) our NETNet on small objects. `FN' represents the missing detection error (false negative). The overall false negative error can be measured by subtracting the AP value of BG from FN. The overall false negative error is 60.8\% for SSD and 39.8\% for NETNet. Lower is better.}
    \label{Error}
\end{figure}

\noindent \textbf{Tackling FN problem.} We show the error analysis plots of our baseline SSD and our NETNet in Fig.~\ref{Error} for small objects. Each plot describes a Precision Recall (PR) curve obtained by eliminating the corresponding detection errors except `C75' (\ie, AP$_{75}$) and `C50'(\ie, AP$_{50}$). Thus, the area of each color can measure the corresponding errors. Overall, our method is more significant on small object detection (\ie, 39.8\% FN error by NETNet \textit{vs} 60.8\% error by SSD). As shown in Fig.~\ref{figurevis}(d), our NETNet can detect small objects precisely, and alleviate the FN problem well.

\section{Conclusion}
In this paper, we have proposed a Neighbor Erasing and Transferring (NET) mechanism with feature reconfiguration for tackling complex scale variations in object detection. Scale-aware features are generated by erasing the features of larger objects from the shallow layers and transferring them into deep pyramid layers. 
We have constructed a single-shot network called NETNet by embedding NETM and NNFM to achieve fast and accurate scale-aware object detection. As demonstrated by experiments on the MS COCO dataset, our NETNet is able to solve the missing detection and part false positive problems effectively, leading to an improved trade-off for real-time and accurate detection. In future work, we consider to explore the advantages of NET on other detectors for scale-aware object detection.


{\small
\bibliographystyle{ieee_fullname}
\bibliography{egbib}
}

\end{document}